\documentclass[twocolumn,10pt]{article}
\usepackage[a4paper,margin=0.6in]{geometry}
\usepackage{cite}
\usepackage{amsmath,amssymb,amsfonts}
\usepackage{algorithmic}
\usepackage{graphicx}
\usepackage{textcomp}
\usepackage{booktabs}
\usepackage{authblk}
\usepackage{hyperref}
\usepackage{caption}
\usepackage{enumitem}
\setlist{nosep}

\usepackage[scaled=0.92]{helvet}

\usepackage{newtxmath} 

\renewcommand{\refname}{\normalsize REFERENCES}


\renewcommand{\thesection}{\Roman{section}}
\renewcommand{\thesubsection}{\Alph{subsection}}
\renewcommand{\thesubsubsection}{\arabic{subsubsection})}

\usepackage{titlesec}
\titleformat{\section}{\normalsize\bfseries\centering}{\thesection.}{0.5em}{\MakeUppercase}
\titleformat{\subsection}{\normalsize\bfseries\itshape}{\thesubsection.}{0.5em}{}
\titleformat{\subsubsection}{\normalsize\bfseries\itshape}{\thesubsubsection}{0.5em}{}
\titlespacing{\section}{0pt}{1.5ex plus 0.5ex}{1ex plus 0.3ex}
\titlespacing{\subsection}{0pt}{1ex plus 0.3ex}{0.5ex plus 0.2ex}
\titlespacing{\subsubsection}{0pt}{0.8ex plus 0.2ex}{0.3ex plus 0.1ex}

\newcommand{\history}[1]{}
\newcommand{\doi}[1]{}

\renewcommand{\markboth}[2]{}

\newcommand{\PARstart}[2]{#1#2}

\newcommand{\authorrefmark}[1]{}
\newcommand{\EOD}{}

\newcommand{\address}[2][]{}

\usepackage{bm}

\def\BibTeX{{\rm B\kern-.05em{\sc i\kern-.025em b}\kern-.08em
    T\kern-.1667em\lower.7ex\hbox{E}\kern-.125emX}}

\makeatletter
\begin{document}
\history{}
\doi{}

\twocolumn[
\begin{@twocolumnfalse}

\title{Physical Imitation Learning: \\Distilling Control Policies into Passive Elasticity}

\author{Huyue Ma, Yurui Jin, Helmut Hauser, and Rui Wu\thanks{Corresponding author: Rui Wu (rui.wu@bristol.ac.uk). This work was supported by the Royal Society Newton International Fellowship NIF/R1/241883, and the Engineering and Physical Sciences Research Council (EPSRC) grant EP/S021795/1.}}
\date{\normalsize School of Engineering Mathematics and Technology, University of Bristol,\\ and the Bristol Robotics Laboratory, BS8 1QU, United Kingdom}

\maketitle

\begin{abstract}
\noindent
Due to brain-body co-evolution, animals' intrinsic body dynamics play a crucial role in their energy-efficient locomotion. Specifically, the control effort is shared between active muscles and passive body dynamics---a principle often referred to as Physical Intelligence. As a result, the body dynamics are part of the solution. In contrast, robot bodies are typically designed to be as simple as possible, but the active control often fights the intrinsic body dynamics, resulting in low energy-efficiency. We introduce Physical Imitation Learning (PIL), a novel approach that brings current robotics control closer to animals. PIL takes learned control policies obtained with Reinforcement Learning (RL) and systematically splits them up into an active and passive control contribution. The passive part can be then directly offloaded to passive Parallel Elastic Joints (PEJs). As a result, the active control contribution is significantly reduced, lowering the overall energy consumption. Furthermore, the policy can be trained via RL to leverage the PEJ assistance by generating gaits that are more readily emulated by the PEJs. This enables co-design of the active and passive control components, shifting a greater share of actuation effort to the PEJs. Here we demonstrate the potential of this approach in simulated quadrupeds. Our results show that the proposed approach can offload up to 95\% of mechanical power to passive body dynamics on flat terrain and 13\% on rough terrain. PIL thereby provides a generalisable route to task-specific Physical Intelligence applicable to a wide range of joint-based robot morphologies.
\end{abstract}

\vspace{1em}

\end{@twocolumnfalse}
]

\section{Introduction}
\label{sec:introduction}

\begin{figure*}[t!]
\centering
\includegraphics[width=0.7\textwidth]{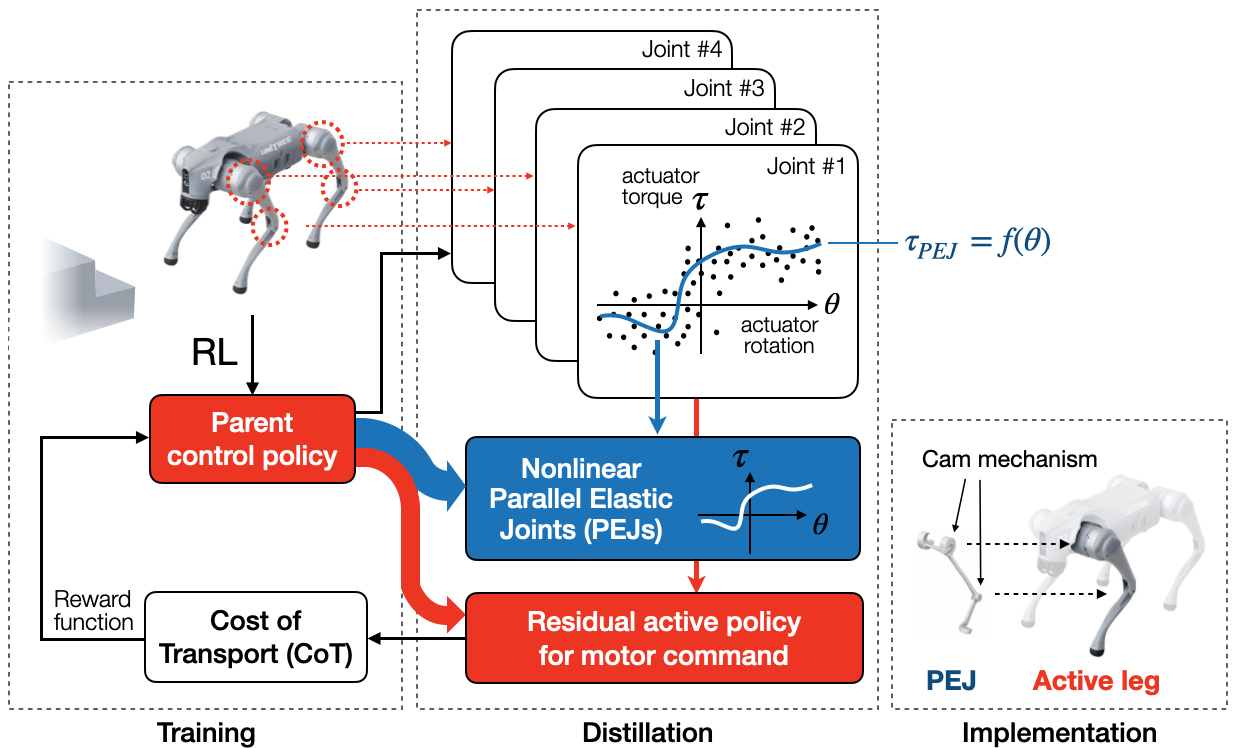}
\caption{\textbf{The Physical Imitation Learning pipeline.} From left to right: the learned control policy is trained via RL with a Cost-of-Transport reward. Actuator torque-angle history recorded from each joint are distilled into nonlinear Parallel Elastic Joint (PEJ) characteristic profiles. The residual torque after PEJ subtraction defines the active policy for motor commands. The PEJs can be physically realised using cam mechanism.}
\label{fig:pipeline}
\end{figure*}

\PARstart{A}{nimals} achieve highly energy-efficient legged locomotion by offloading part of their control effort to the passive dynamics of their bodies--a phenomenon often referred to as Physical Intelligence~\cite{burden2024animals,hauser2024morphological,kashiri2018overview}. The intrinsic passive mechanics of the limbs help shaping robust gaits, while muscles activities provide the necessary actuation and control to maintain and modulate this motion. Animal locomotion is therefore a complex interplay between active neuromuscular activities and passive body dynamics~\cite{kuo2010dynamic,daley2009role}. This is a natural consequence of brain-body co-evolution that fine-tuned the nonlinear compliance of the musculoskeletal system to store elastic energy and release it in synchrony with muscle actuation~\cite{roberts1997muscular,biewener2000muscle,ishikawa2005muscle,hauser2023leveraging}. This highly effective exploitation of passive dynamics (Physical Intelligence) has been identified as one of the key aspects to animals superior energy efficiency over legged robotic systems~\cite{pfeifer2006body,burden2024animals,kashiri2018overview}. Currently, legged robots typically rely on a centralised control approach, where body dynamics are not designed to contribute to the control task and, moreover, often even fight against the controller. As a result, today’s legged robots still exhibit substantially higher energy demand compare to their biological counterparts. The scale of this gap can be quantified by the dimensionless Cost of Transport (CoT), i.e., energy required per unit weight per unit distance. Modern legged robots have a CoT of $\approx$ 0.5–1.5~\cite{seok2013design,hutter2016anymal} compared to CoT $\approx$ 0.2–0.5 achieved by comparable terrestrial mammals ~\cite{taylor1982energetics}.
\\
However, currently, Physical Intelligence in robots is mostly driven by intuition and/or bio-inspiration. It lacks a general and rigorous framework. Famous examples are Passive Dynamic Walkers which are capable of stable bipedal walking without active control. This is achieved by offloading the stabilisation control to the passive dynamic response of a carefully tuned mechanical body~\cite{collins2005efficient}. However, they are fine-tuned to a specific slope and small environmental changes, e.g., a change of slope angle renders them useless. Based on the same mechanical principle, the Cornell Walker has achieved human-level energy efficiency~\cite{bhounsule2014low}, but only on flat ground and very constrained conditions. It lacks the robustness required for more unstructured environments. Bio-inspiration has been another approach to implement Physical Intelligence. Researchers have built robot legs that replicate the tendon-coordinated articulation of animals to offload low-level control to the body. The results have shown self-regulated motion and high energy efficiency~\cite{cotton2012fastrunner,badri2022birdbot,lin2025bird}. Bio-inspiration has also realised Physical Intelligence at full-body level, where motion patterns were extracted from animal locomotion and embedded into optimised mechanical bodies. This enabled the robot body to physically mimic animal body movements~\cite{ramezani2017biomimetic,stella2025synergy}. Beyond intuition and bio-inspiration, data-driven and learning-based co-design approaches have begun to explore brain–body co-evolution to jointly optimise body and control as well~\cite{bjelonic2023learning,luck2020data,cheney2018scalable}. This allows Physical Intelligence to be designed for specific control tasks beyond those encountered by animals or achieved by manually tuned mechanisms. However, artificial co-evolution typically requires searching simultaneously over both the controller and morphological parameter space. The resulting high-dimensional optimisation problem is susceptible to the curse of dimensionality, and scaling such approaches efficiently remains an open challenge~\cite{burden2024animals,goff2024investigation,nagiredla2024ecode}. Therefore, the field needs a computationally-lightweight, generalisable design framework that can derive body designs from task-specific target control policies. 
\\
Therefore, we introduce Physical Imitation Learning (PIL), a design pipeline that uses learned RL control policies and splits them up into a passive and an active component, see Fig.~\ref{fig:pipeline}. The passive part can be outsourced and implemented directly in nonlinear Parallel Elastic Joints (PEJs), for example, in form of an Elastic Rolling Cam's as introduced in~\cite{wu2025encoding}. The active part is retained in the central controller. Through the systematic outsourcing of parts of the control policy, the overall energetic effort can be reduced significantly since only the now-reduced residual control policy requires energy. To our knowledge, this is the first rigorous framework that allows to distil passive components of a control policy directly into physically feasible passive components.
The distillation process takes in the controller-generated actuation history during simulated operation on various terrains, and converts it into nonlinear torque--angle responses of the PEJs to mechanically emulate the actuation history.
During operation of a robot equipped with such PEJs, the PEJs generate passive torques to assist motor actuation, while the motor actuation naturally synchronises with the PEJs--because they are complementary components of the same policy--to fully reproduce the learned policy's behaviour. 
Furthermore, co-design of the policy and the PEJ profiles can be achieved by training the policy to leverage the PEJ assistance, converging towards gaits whose torque demands are more readily emulated by the passive torque--angle profile.
In the present study, we demonstrate that PIL can offload up to 95\% of the mechanical power to PEJs for a simulated Unitree Go2 quadruped robot. PIL provides the first generalisable framework that realise task-specific Physical Intelligence that use PEJ to physically imitate and thereby assist a RL control policy, which can be applied to a wide range of joint-based robot morphologies.
In the next section we will provide an outline of the proposed pipeline. After that, we detail its three components---the parallel elastic joints, the distillation procedure, and the two-stage training protocol---and then present simulation results quantifying the energy offloaded to the passive elements across a range of terrains. We further show that the distilled elastic profiles are physically manufacturable.

\section{Physical Imitation Learning}
\label{sec:pipeline}

This section provides an overview of the PIL pipeline (Section~\ref{sec:PIL}), then details the PEJ (Section~\ref{sec:PEJ}), the distillation procedure (Section~\ref{sec:distillation}), and the two-stage training protocol used in this study that co-designed the policy with the PEJ (Section~\ref{sec:training}).

\subsection{Overview of the PIL Pipeline}
\label{sec:PIL}

The PIL pipeline (Fig.~\ref{fig:pipeline}) takes three steps. First, the training of control policy via standard Reinforcement Learning (RL). Second, splitting of this policy through a distillation process, which converts the leg's actuation history collected during simulated locomotion into the nonlinear elastic response of Parallel Elastic Joints (PEJs)---spring-like elements whose torque depends only on joint angle. Third, the PEJ is placed in parallel with the motor, so that during operation it supplies passive torque based on the joint angle, thus the motor only needs to provide the residual active torque. Although the present study is based on simulation only, the manufacturability of the required PEJs was verified by previous study~\cite{wu2025encoding}, and further discussed in Section~\ref{sec:PEJ_feasibility}. The result is that a large fraction of the actuation effort can be offloaded from the battery-powered motor to the passive PEJ, without changing the robot's behaviour.

\subsection{Parallel Elastic Joints}
\label{sec:PEJ}
A Parallel Elastic Joint (PEJ) required for the pipeline is a passive mechanical element that sits in parallel with the motor and produces a restorative torque $\tau_{PEJ}$, which depends only on the current joint angle $q$. This means it is a rotational spring with a nonlinear stiffness function. We refer to this $\tau_{\text{PEJ}}=f(\theta)$ as the the profile. While we only show simulation results here, note that such profile can be physically realised, e.g., using the Elastic Rolling Cam mechanism introduced in~\cite{wu2025encoding}. The PEJ delivers its torque passively as the joint rotates through each angle, and consumes no energy of its own, i.e, only stores and releases energy. The PEJ produces therefore part of the torque that the motor would otherwise have to produce if no passive component would be available, thereby reducing the overall motor effort and, hence, battery energy consumption. The residual torque, i.e., the difference between the total joint torque demanded by the control policy and the PEJ torque (which we refer here to passive control), is supplied by the motor (which we refer here to active control).
The next section describes how the nonlinear profile can be extracted from the RL control policy.

\subsection{Distillation}
\label{sec:distillation}

\begin{figure*}[t!]
\centering
\includegraphics[width=\textwidth]{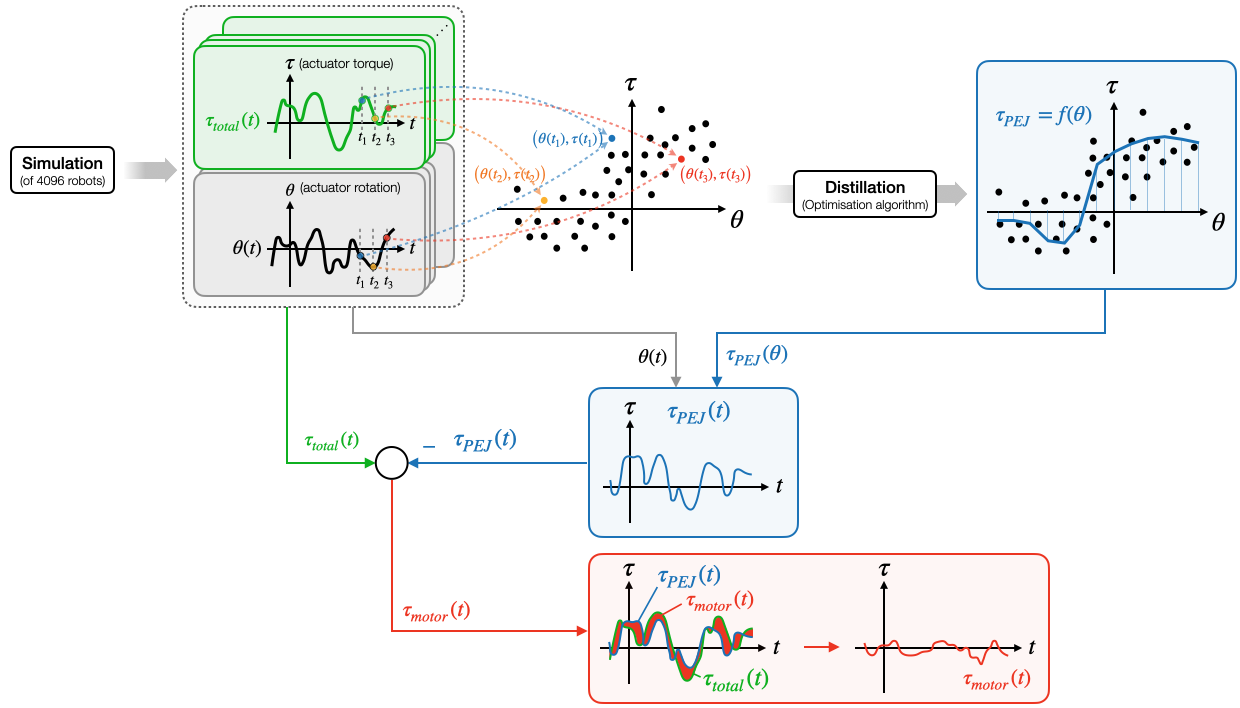}
\caption{\textbf{Distillation procedure.} The joint torque and angle time series are recorded from the parallel simulation of 4,096 robots walking under various speed commands. Top: these time series are mapped to a torque--angle scatter plot, and a piecewise-linear PEJ profile is fitted via bin-wise optimisation to minimise the residual motor power consumption, which is physically implementable using a cam mechanism. Bottom: at each time step, the PEJ torque $\tau_{\text{PEJ}}(t)$ is subtracted from the total demanded torque $\tau_{\text{total}}(t)$, leaving a residual motor torque $\tau_{\text{residual}}(t)$.}
\label{fig:distillation}
\end{figure*}

The distillation stage extracts systematically the passive component of the learned policy and encodes it into nonlinear PEJ profiles, as illustrated in Fig.~\ref{fig:distillation}. The procedure works as follows. When the learned RL control policy is employed, each joint goes through a torque--angle time series (i.e., the actuation history) as a results of the RL controller commands. These time series are collected, and recast as a relationship between joint angle $\theta$ and torque $\tau$ (Fig.~\ref{fig:distillation}, top). The points are scattered because they are taken from a total of 4,096 robots under various speed commands in simulation (see for more details in Section~\ref{sec:training}). However, a PEJ profile, being a passive response only, can only be represented by a fixed continuos function, i.e., a single profile. The distillation process is set up to find the one optimal profile $\tau_{\text{PEJ}}= f(\theta)$, so that when its torque is subtracted from the demanded torque by the RL policy, it removes as much of the motor's energy consumption as possible (Fig.~\ref{fig:distillation}, bottom). What remains after the subtraction of PEJ torque is the residual motor torque, or the active control component. Note that the more periodic a gait is, the tighter the scatter clusters around a single profile, and the more of the actuation effort the PEJ can replace, i.e., outsourced to the passive component. This means for more complex terrains it becomes hard to find a good single profile.

The goals is to find for each joint $j$ a corresponding torque--angle profile $\tau_{\text{PEJ}}=f(\theta)$ that outsources as much of the actuator effort as possible, leaving only a residual torque for the motor. With the PEJ assistance, the total torque commanded by the learned policy can then be decomposed into
\begin{equation}\label{eq:torque_decomposition}
\tau_{\text{total}} = \tau_{\text{PEJ}} + \tau_{\text{residual}},
\end{equation}
so that the motor's mechanical power becomes
\begin{equation}\label{eq:motor_power}
P_{\text{mechanical}} = \bigl(\tau_{\text{total}} - \tau_{\text{PEJ}}\bigr)\,\dot{\theta}.
\end{equation}

For each $\text{PEJ}$, the nonlinear torque--angle profile $\tau_{\text{PEJ}}=f(\theta)$ is a piecewise-linear function, which can easily be implemented physically using a cam mechanism. The piecewise-linear function is defined by $20$ equally spaced points connected by line segments. The torque value of the $20$ points $\{\tau_{\text{PEJ}}(\theta_j)\}_{j=1}^{20}$ are optimised from the joint--angle actuation history, which is divided into $20$ equally-spaced bins across the joint angle. Let $T_b$ denote the set of simulation time steps at which the joint angle falls within bin $b$. For each bin we seek a single constant PEJ torque $\tau_{\text{PEJ}}^*(\theta_b)$ that minimises the total motor power consumption summed over all those time steps sampled from all the $4096$ simulated robots:
\begin{equation}\label{eq:pea_optimisation}
\tau^{*}_{\text{PEJ}}(\theta_b) = \arg\min_{\tau_{\text{PEJ}}} \sum_{t \in T_b} P(t),
\end{equation}
where the \emph{motor power consumption} $P$ is the positive mechanical power the motor actively delivers to the joint. This also means at time steps when the motor power is negative (i.e., when the motor would dissipate energy rather than deliver it), $P$ is set to zero under a zero-regeneration assumption. As explained later in Section~\ref{sec:power}, this provides a realistic approximation of motor power drawn from the battery.

The $20$ per-bin torques $\{\tau_{\text{PEJ}}(\theta_j)\}_{j=1}^N$ are optimised jointly by gradient descent over the total motor power. Throughout RL training we maintain a rolling experience buffer that stores the $200{,}000$ most recent joint-data tuples $(\theta(t),\,\dot{\theta}(t),\,\tau_{\text{total}}(t))$ per joint type (i.e., front calf, rear calf, front thigh, and rear thigh, as detailed later in Section~\ref{sec:training}) at a sample rate of $50$\,Hz and across all the $4096$ simulated robots. At each gradient-descent run, a minibatch of 20\% of the buffer (up to $40,000$ samples) is evenly drawn from the buffer; for every sample in the minibatch, the motor power consumption under the current PEJ profile is evaluated from Eq.~\eqref{eq:motor_power} (negative values set to zero; see Section~\ref{sec:power}). Samples whose joint angles lie in the lowest or highest $5\%$ of the observed angle distribution are excluded from the update. This exclusion process reduced the PEJ angle range, so that the $20$ bins are focused on the most frequently used angle range. The optimisation algorithm updates the torques per bin to minimise the summed objective in Eq.~\eqref{eq:pea_optimisation}. After convergence, the $20$ per-bin torque values are connected by straight line segments to form a piecewise-linear PEJ profile $\tau_{\text{PEJ}}(q)$. Outside the bin range, the profile is linearly ramped to zero over a 5-degree margin on either side, and remains zero beyond. While we use only simulation here, this piecewise-linear PEJ profile is physically implementable using cam mechanism, which is further discussed in Section~\ref{sec:PEJ_feasibility}.

Note that, the control policy is trained to generate the total torque $\tau_{\text{total}}$, which is then separated into $\tau_{\text{PEJ}}$ and $\tau_{\text{residual}}$ (according to Eq.~\ref{eq:torque_decomposition}) through distillation that does not require explicit simulation of the PEJs in the training environment. This means that the RL training loop only needs to optimise the control policy---same as in conventional RL---without adding any mechanical design parameters to the search space. The dimensionality of the search space therefore remains manageable, and the pipeline avoids the curse of dimensionality that typically afflicts joint brain-body co-design. In PIL, co-design of the policy and the PEJ profiles is achievable because the policy can be trained to increase the energy-efficiency term in the reward function by leveraging the PEJ assistance---converging toward gaits whose torque demands are more readily replaced by the passive torque--angle profile, which is further explained in Section~\ref{sec:training}.

\subsection{Training Protocol}
\label{sec:training}

This section describes the training protocol used. Training proceeds in two stages: (1) pre-training a policy via curriculum learning, and, (2), continued training that adds an energy-efficiency penalty to the reward together with online PEJ distillation to realise co-design of the controller (i.e., the active control component) and the PEJ (i.e., the passive control component). The purpose of Stage~1 is to obtain a locomotion policy that walks robustly across the full terrain curriculum. This pre-trained policy then serves as the shared starting point for all of the Stage~2 experiments, to keep the overall computational requirements manageable. The details of both step are described in the following sections.

All experiments were conducted in NVIDIA Isaac Sim 2023.1.1 using the Orbit framework (v0.3.0), which provides rigid-body physics and GPU-accelerated parallel execution~\cite{mittal2023orbit}. Policies were trained with PPO via the integrated RSL-RL library. We simulated 4096 instances of the Unitree Go2 quadruped robot in parallel, at a control frequency of $50$\,Hz (i.e., $\Delta t{=}0.02$\,s). Training uses IsaacLab's default terrains, which defines seven discrete difficulty levels (flat terrain and Levels~1--6; see Fig.~\ref{fig:combined_results}(c) and Appendix Fig.~\ref{fig:terrains} for visual examples). Each level is a mixture of five procedurally generated terrain primitives: flat ground; random rough surfaces (bumpy ground with many small, irregular height variations, resembling unpaved natural terrain); box obstacles; pyramid stairs (ascending staircases); and inverted pyramid stairs (descending staircases). As the level increases, a larger fraction of each patch is made up of the more demanding primitives and their dimensions grow: flat terrain is entirely flat, whereas Level~6 combines random roughness with amplitudes up to $7.3$\,cm, box obstacles up to $15$\,cm tall, and stair risers up to $17$\,cm (full per-level specifications in Table~\ref{tab:terrain_specs} in Appendix).

The Go2 model has a total of 12 actuated joints (three per leg, named here following the Unitree/IsaacLab URDF convention: \emph{hip} for hip abduction/adduction, \emph{thigh} for hip flexion/extension, and \emph{calf} for knee flexion/extension), and are parameterised with the manufacturer-specified mass distribution, joint limits, and actuator dynamics, including a maximum motor torque of $23.5$\,Nm per joint. PEJs are added only to the joints with the highest energy consumption, i.e., the four thigh and four calf joints, and therefore excludes the four hip joints, whose power draw is comparatively low (over $5\times$ lower than that of the thigh and calf joints in the final trained policy). Exploiting the bilateral symmetry of the quadruped, the eight target joints are grouped into four symmetric pairs (front thigh, rear thigh, front calf, rear calf), and the two joints within each pair use PEJs with the same torque--angle profile. 

The RL policy is implemented as a neural network that maps observations to target joint positions. The observations comprise all joint positions and velocities of all 12 joints, the base orientation and base angular velocity (i.e., the tilt of the robot's torso and its angular rate of rotation, which together tell the policy how the body is oriented and moving), and a height scan of the local terrain (a grid of ground-height samples around the robot that conveys upcoming obstacles). The network is a multi-layer perceptron with three hidden layers of 512, 256, and 128 units and Exponential Linear Unit (ELU)    activations.

\subsubsection{Stage~1: Pre-training}

As mentioned before, Stage 1 is meant to provide stable starting point for Stage 2. The policy was trained from scratch using Proximal Policy Optimisation (PPO)~\cite{schulman2017proximal} via the RSL-RL framework~\cite{rudin2022learning}. Training ran in $4096$ parallel environments (i.e., with 4096 parallel robots). At the start of each episode, each robot was assigned a random target consisting of forward speed, lateral speed, and yaw rate, sampled uniformly from $[0, 2]$\,m/s, $[-0.5, 0.5]$\,m/s, and $[-1, 1]$\,rad/s, respectively. Note that the lateral and yaw velocities were included to ensure a generalisable gait. 
The following default terrain curriculum is applied during pre-training: each robot starts on flat terrain and is promoted to the next difficulty level as soon as, within a single episode, it has walked more than half of the distance that a perfectly-tracking agent would cover at its target velocity over the full episode.

The reward at each control step, $r_{\text{base}}$, is a weighted sum of individual components,
\begin{equation}\label{eq:base_reward}
r_{\text{base}} = \sum_{k} w_k\, r_k,
\end{equation}
where each $r_k$ is an individual reward term and $w_k$ its corresponding weight (listed in full in the Appendix, Table~\ref{tab:reward_weights}). The two dominant positive components reward \emph{linear-velocity tracking} (how well the robot matches its target forward speed) and \emph{angular-velocity tracking} (how well it matches the commanded yaw rate), both using an exponential kernel that equals $1$ at perfect tracking and decays smoothly as the error grows. The remaining components reward body stability (maintaining an upright posture and standing height) and penalise undesired behaviour such as excessive joint torques, large joint accelerations, and action jerk. The base-height term is a one-sided penalty that prevents the robot from adopting an energy-saving but undesirable crawling gait. 

The policy is trained for $10{,}000$ iterations. Convergence is observed after approximately $6{,}000$ iterations, at which point the curriculum has plateaued at the highest achievable level, i.e., Level~6.

\subsubsection{Stage~2: Co-design of Controller and PEJ}
Stage~2 continues from the pre-trained policy from Stage 1 with an added \emph{Cost-of-Transport} (CoT) penalty that targets energy efficiency. CoT is the mechanical energy a robot consumes to travel one unit of distance, normalised by its weight---a dimensionless measure of how efficiently a given gait converts energy into forward motion. As described in Section~\ref{sec:distillation}, the four hip joints are excluded because their power draw is over $5\times$ lower than that of the thigh and calf joints. The CoT is computed as
\begin{equation}\label{eq:cot}
\text{CoT} = \frac{\displaystyle\sum_{i \in \{\text{thigh},\,\text{calf}\}} P_i(t)}{m\,g\,v_{\text{scalar}}},
\end{equation}
where $P_i(t)$ is the motor power consumption at joint $i$, $m{=}15$\,kg is the robot mass, and $g{=}9.81$\,m/s$^2$. When the projected speed drops below $0.1$\,m/s, we clamp $v_{\text{scalar}}$ to $0.1$\,m/s, which prevents the CoT from blowing up at near-stationary movement. Because CoT divides energy by distance, a policy could artificially lower it by steering around obstacles. To prevent this, the velocity used in the CoT denominator is a projected forward speed,
\begin{equation}\label{eq:v_scalar}
v_{\text{scalar}} = \max\,\bigl(0,\;\vec{v}_{\text{actual}}\cdot\hat{v}_{\text{cmd}}\bigr),
\end{equation}
where $\vec{v}_{\text{actual}}$ is the robot's current velocity vector and $\hat{v}_{\text{cmd}}$ is the unit vector along the desired (target) direction. To suppress sudden velocity fluctuations, $v_{\text{scalar}}$ is averaged over a sliding window of $10$ simulation time steps, which was chosen empirically.

The training reward in Stage~2 becomes
\begin{equation}\label{eq:cot_reward}
r_{\text{total}} = r_{\text{base}} - \alpha\,\text{CoT},
\end{equation}
where $\alpha\geq 0$ is the CoT weight and was set according to the terrain complexity as discussed later. During Stage~2 and all subsequent evaluations, the same lateral-speed and yaw-rate ranges as in pre-training were used ($[-0.5,0.5]$\,m/s and $[-1,1]$\,rad/s), while the forward speed was sampled from $[0.5, 2]$\,m/s---narrowed from the pre-training range of $[0,2]$\,m/s to exclude near-stationary commands for which the CoT becomes ill-conditioned.

A large $\alpha$ reduces CoT at the expense of locomotion quality. On flat terrain the gait is more periodic, so a large $\alpha$ can be applied before tracking accuracy degrades (echoing the principle underlying passive dynamic walkers~\cite{collins2005efficient}). On rough terrain the gait become more non-periodic, and the control must be more active, which limits the values for $\alpha$. We therefore tune $\alpha$ separately for each target terrain level. Starting from the pre-trained policy, we train the policy for $10{,}000$ iterations at a candidate $\alpha$, measure the resulting mean absolute velocity-tracking error (Eq.~\eqref{eq:tracking_error}), and compare it to that of the pre-trained policy (${\approx}0.1$\,m/s across terrain levels). We adjust $\alpha$ and re-train the policy so long as the trained policy's tracking error stays within $\pm0.01$\,m/s from the pre-trained reference. The resulting per-terrain values are listed in the Appendix, Table~\ref{tab:cot_weights}.

The distillation procedure of Section~\ref{sec:distillation} runs online: every $24$ environment steps---i.e., once per policy rollout---the PEJ profiles are re-fitted to convergence on the current experience buffer, using a gradient-descent inner loop with classical (heavy-ball) momentum, where each step updates the per-bin torques in the direction of the negative gradient of the total motor power, blended with a fraction of the previous step's update. We use a learning rate of $0.15$ and a momentum coefficient of $0.8$, which accelerates convergence and smooths the update across the noisy, scatter-distributed joint data.
Because the PEJ torque is already subtracted from the motor torque at every simulation step, a portion of the motor power is reduced by the PEJ before the CoT penalty is evaluated. This enables co-design of the policy and the PEJ profiles, as the policy learns to leverage the PEJ assistance by converging toward gaits whose torque demands are more readily replaced by the passive torque--angle profile.

To verify the effect of the co-design and to show that the energy savings come from the PEJs rather than from the CoT reward alone, we also train a \emph{reference} policy for each terrain level which is identical to the co-design policy in every respect but \emph{without} PEJ distillation. Its CoT weight, $\alpha$ is tuned by the same procedure described above, and the resulting values are also listed in Table~\ref{tab:cot_weights}. The CoT weights of the reference policy are smaller than those of the co-design policy, since without PEJ assistance, a large $\alpha$ will cause significant tracking-error because a substantial CoT reduction is not achievable without significantly reducing the locomotion performance. 


\section{Results}
\label{sec:results}
We evaluated the PIL pipeline on the simulated Unitree Go2 robot. Section~\ref{sec:terrain_results} reports the energy savings and distilled PEJ profiles of the co-design policies trained on each of the 6 individual terrain level. Section~\ref{sec:cross_terrain} presents the results of the tests on how well these terrain-specific policies generalised to other terrains.
\subsection{Terrain-Specific Power Reduction}
\label{sec:terrain_results}
\begin{figure*}[t!]
\centering
\includegraphics[width=0.95\textwidth]{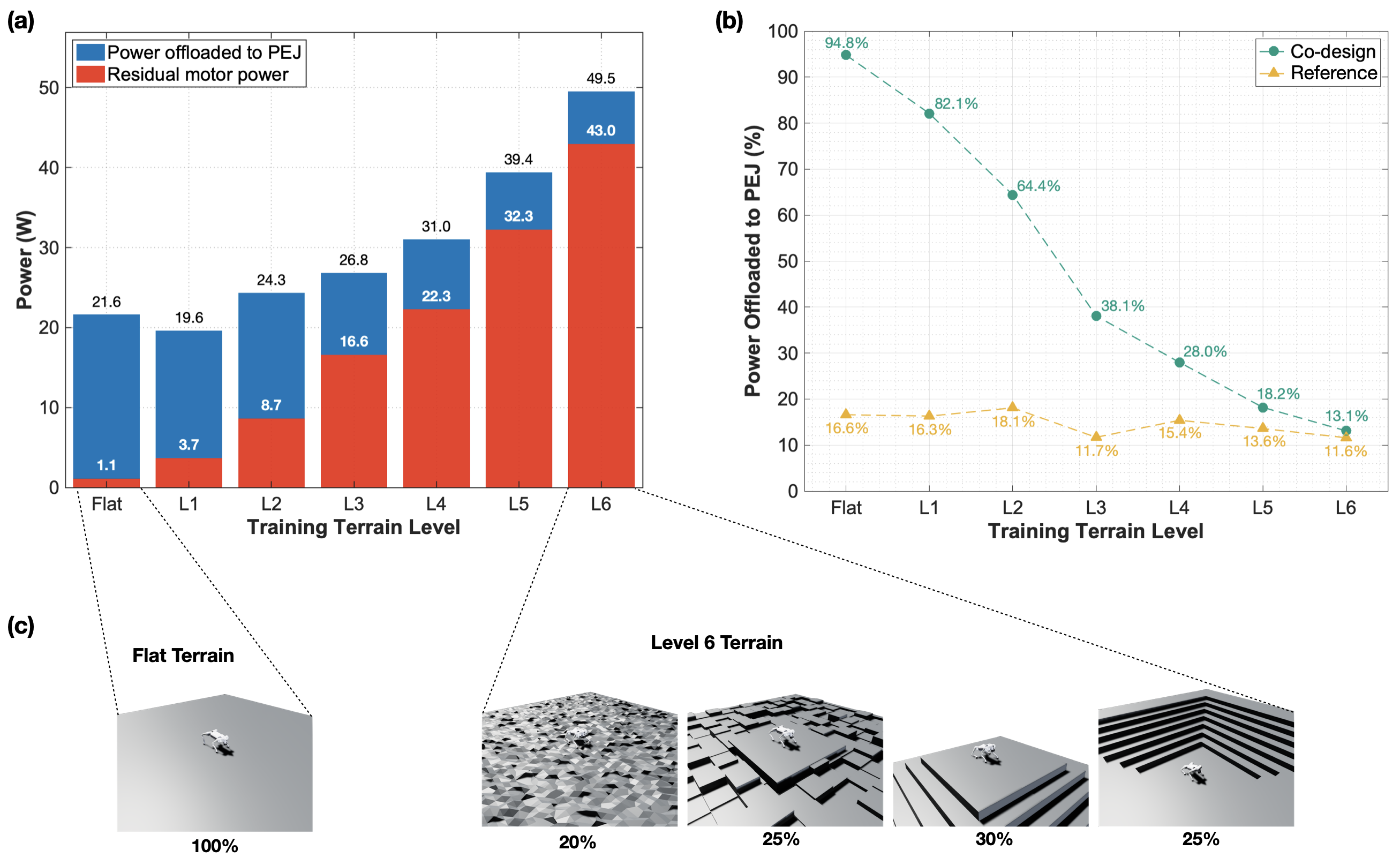}
\caption{\textbf{Terrain-specific power reduction.} (a) Power decomposition of each terrain-specific co-design policy. The stacked bars show total power consumption split into the active motor power (red) and the PEJ contribution (blue), the consumption include the hip motors. (b) Offload percentage versus training terrain level. Green circles: co-design policy (with PEJ vs.\ without PEJ). Yellow triangles: offload percentage obtained by fitting a PEJ profile post-hoc to the reference policy. (c) Visual examples of flat terrain and level~6 terrain, showing the terrain primitives and their proportions, the full terrain composition is shown in Fig.~\ref{fig:terrains}.}
\label{fig:combined_results}
\end{figure*}

We first investigated how the total actuation effort split between the active motors and the passive PEJs, and how that split changed with terrain difficulty. Fig.~\ref{fig:combined_results}(a) decomposes the average motor power consumption of each terrain-specific co-design policy into the active motor contribution (red) and the PEJ contribution (blue).

We quantified the PEJ assistance by the \emph{power offload percentage}, i.e., the fraction of the co-design policy's actuation effort that was replaced by the PEJs,
\begin{equation}\label{eq:offload_ratio}
R_{\text{offload}} = \frac{\bar{P}_{\text{w/o\,PEJ}} - \bar{P}_{\text{w/\,PEJ}}}{\bar{P}_{\text{w/o\,PEJ}}} \times 100\%,
\end{equation}
where $\bar{P}_{\text{w/\,PEJ}}$ is the average motor power consumption (Eq.~\eqref{eq:positive_power}) of the co-design policy with PEJ assistance active, and $\bar{P}_{\text{w/o\,PEJ}}$ is that of the same policy without the PEJ torque (i.e., the motor must supply the full demanded torque).

On flat terrain, the PEJs replaced the vast majority of the actuation effort ($20.6$\,W out of $21.7$\,W, or $95\%$), requiring only $1.1$\,W for the motors (including the consumption of the hip motors). This is because flat-terrain gaits are highly periodic, thus the torque demanded at a given joint angle is nearly the same from one cycle to the next and is therefore well captured by a static angle--torque profile. Consequently, as terrain difficulty increases, the offload percentage $R_{offload}$ decreases monotonically---falling to $13\%$ at Level~6---reflecting the growing need for reactive, non-periodic actuation that cannot be captured by a single torque--angle profile. Therefore, on rough (unpredictable) terrain the policy thereby must keep a larger actuation margin. Throughout, locomotion quality is preserved: the velocity tracking error of each co-design policy stays close to that of the pre-trained policy (the bordered diagonal cells in Fig.~\ref{fig:tracking_heatmap}). The full per-terrain power values for both policies are given in Appendix~\ref{sec:appendix_power}, Table~\ref{tab:power_data}.

Fig.~\ref{fig:combined_results}(b) reports the offload percentage of each co-design policy for terrain level it was trained for (green). To verify the effect of the co-design, we also apply the same PEJ distillation \emph{post-hoc} to the reference policy, i.e., which was trained with a CoT penalty but without PEJ co-design (Section~II\ref{sec:training}). The corresponding offload percentages are shown in Fig. ~\ref{fig:combined_results}(b) in yellow. The gap between the two profiles (co-design and reference) therefore isolates the contribution of co-design from that of the CoT reward. On flat terrain, the co-design policy reaches over $95\%$ offload, whereas the reference policy yields only $16.6\%$. Importantly, the total power consumption of the co-design and reference policies is similar (within $4$\,W at every terrain level), thus the energy saving of the co-design policy remains valid when comparing with the reference policy result.

\begin{figure}[t!]
\centering
\includegraphics[width=\columnwidth]{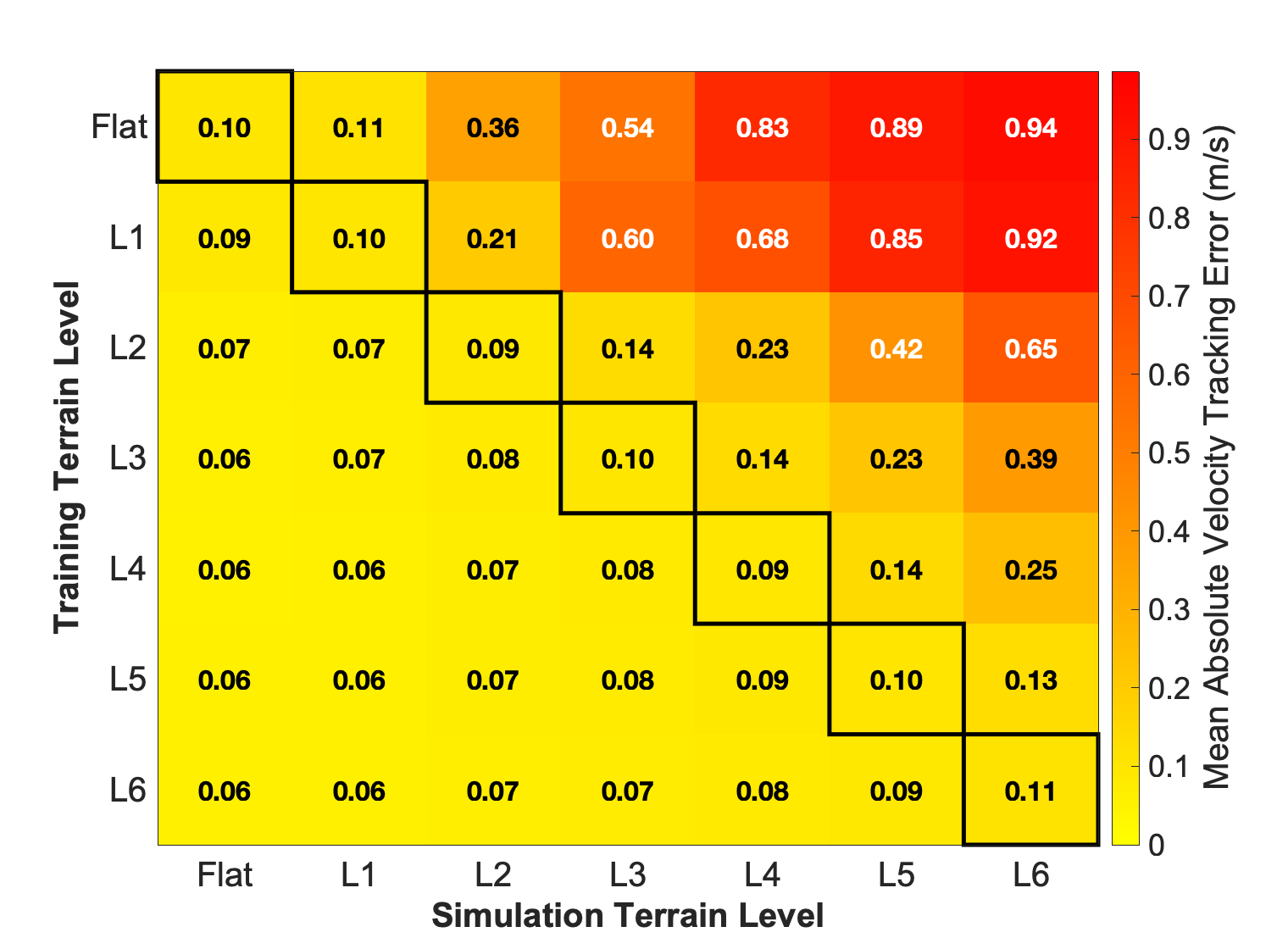}
\caption{\textbf{Mean absolute velocity tracking error heatmap.} Each cell shows the tracking error when deploying a co-design policy trained on the terrain indicated by the row (P$x$) and evaluated on the terrain indicated by the column (T$x$). Bordered diagonal cells correspond to each policy evaluated on its own training terrain.}
\label{fig:tracking_heatmap}
\end{figure}

\begin{figure*}[t!]
\centering
\includegraphics[width=\textwidth]{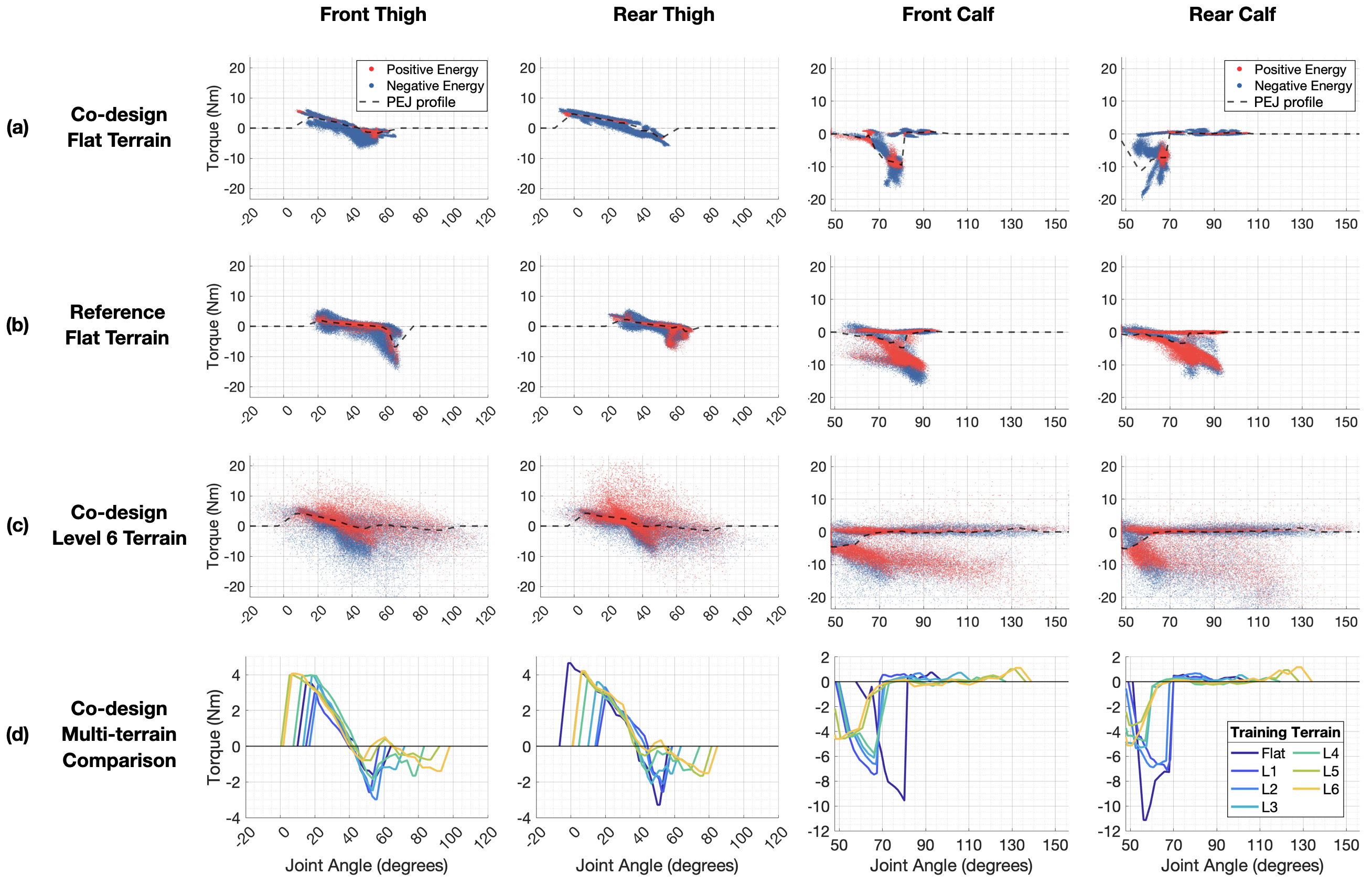}
\caption{\textbf{Distilled PEJ profiles.} Joint names follow the Unitree/IsaacLab URDF convention: \emph{thigh} denotes the hip flexion/extension joint and \emph{calf} denotes the knee flexion/extension joint. (a)--(c) PEJ profiles (dashed lines) for the four symmetric joint groups (columns), overlaid on motor torque--angle data points (dots) recorded from the learned policy during locomotion, with $50{,}000$ data points each. (a) and (b): co-design and reference policies on flat terrain. (c): co-design policy on level~6 terrain. Tighter clustering of the data points around the profile indicates a more periodic gait and greater passive offloading. (d): co-design PEJ profiles overlaid for all seven terrain levels.}
\label{fig:pej_curves}
\end{figure*}

Fig.~\ref{fig:pej_curves} presents the distilled PEJ characteristic profiles for all four joint groups (columns: front thigh, rear thigh, front calf, rear calf), recall that each group pools the left and right joints, which share one profile by bilateral symmetry (Section~\ref{sec:training}). Joint angles follow the convention of the official IsaacLab Go2 configuration, in which zero corresponds to the fully extended leg posture defined in the URDF. The thigh joints therefore operate in positive angles and the calf joints in negative angles during locomotion. For readability of the figures, the calf joint angle axis has been inverted so that all profiles are displayed with positive angles. The dashed lines in the figures are the distilled PEJ torque--angle profiles, and the dots are the motor torque--angle data points recorded from the learned policy during locomotion.

As visible from Fig.~\ref{fig:pej_curves}, the data points with positive energy are clustered more tightly around the PEJ profile under co-design (Fig.~\ref{fig:pej_curves}~(a)) than under the reference policy (Fig.~\ref{fig:pej_curves}~(b)), and more tightly on flat terrain (Fig.~\ref{fig:pej_curves}~(a)) than on Level~6 terrain (Fig.~\ref{fig:pej_curves}~(c)), indicating that co-designing on easier terrain produces the most periodic gaits and the highest proportion of actuation offloaded to the PEJ---consistent with the declining offload percentage in Fig.~\ref{fig:combined_results}. 
Meanwhile, the co-design PEJ profiles from all terrain levels in Fig.~\ref{fig:pej_curves}~(d) show broadly similar shapes, with the primary differences being a wider angular range at more difficult terrains.

\subsection{Cross-Terrain Generalisation}
\label{sec:cross_terrain}
\begin{figure}[t!]
\centering
\includegraphics[width=\columnwidth]{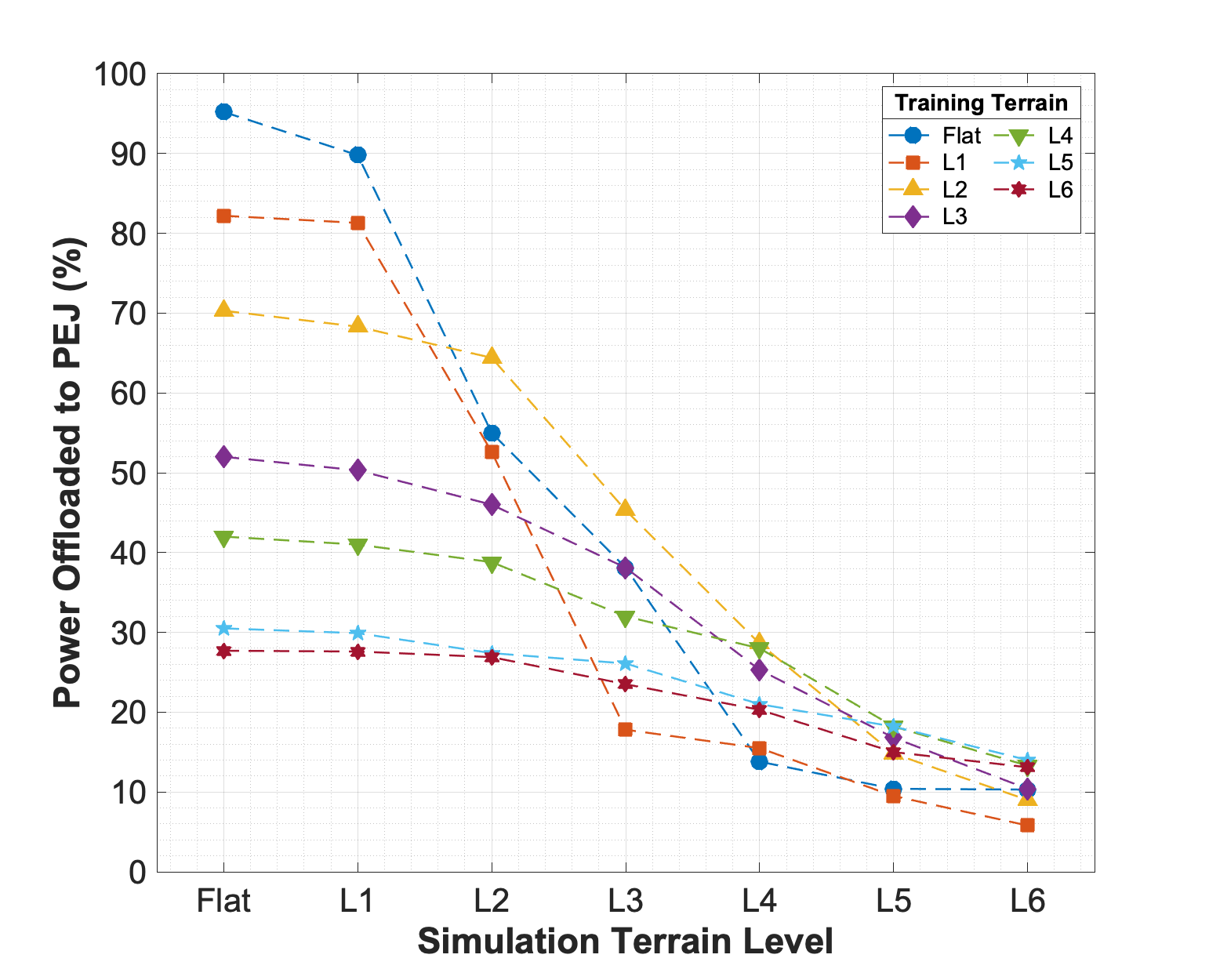}
\caption{\textbf{Cross-terrain power offload percentage.} Each profile represents a co-design policy trained on a specific terrain level and evaluated across all terrain levels. Policies trained on easier terrain achieve higher offload percentages but degrade faster on harder terrain.}
\label{fig:cross_terrain}
\end{figure}

To assess how well each terrain-specific policy generalises beyond its training condition (i.e., the terrain level it was trained on), we deployed each co-design policy on all other terrain levels and measured the resulting power offload percentage. We refer to each policy by its training level (e.g., the flat-terrain policy is trained on flat terrain). Fig.~\ref{fig:cross_terrain} summarises the results.

Interestingly, every policy achieved its highest offload percentage not on their own terrain, but on flat terrain and degrades monotonically with increasing difficulty. At any given evaluation terrain, a policy trained on an easier level usually achieves a higher offload percentage than one trained on a harder level. For example, on flat terrain the flat-terrain policy achieves $95\%$ versus $28\%$ for the Level-6 policy. This because the flat-terrain policy was trained with a higher $\alpha$, which led to more aggressive offloading to the PEJs, but at the cost of reduced tracking accuracy on difficult terrain. The tracking error heatmap (Fig.~\ref{fig:tracking_heatmap}) confirms this trade-off: policies trained on easy terrain exhibit substantially higher tracking error when deployed on hard terrain, whereas policies trained on hard terrain maintain low tracking error across all levels.

\section{Feasibility of manufacturable PEJ}
 \label{sec:PEJ_feasibility}
 
\begin{figure*}[t!]
\centering
\includegraphics[width=0.95\textwidth]{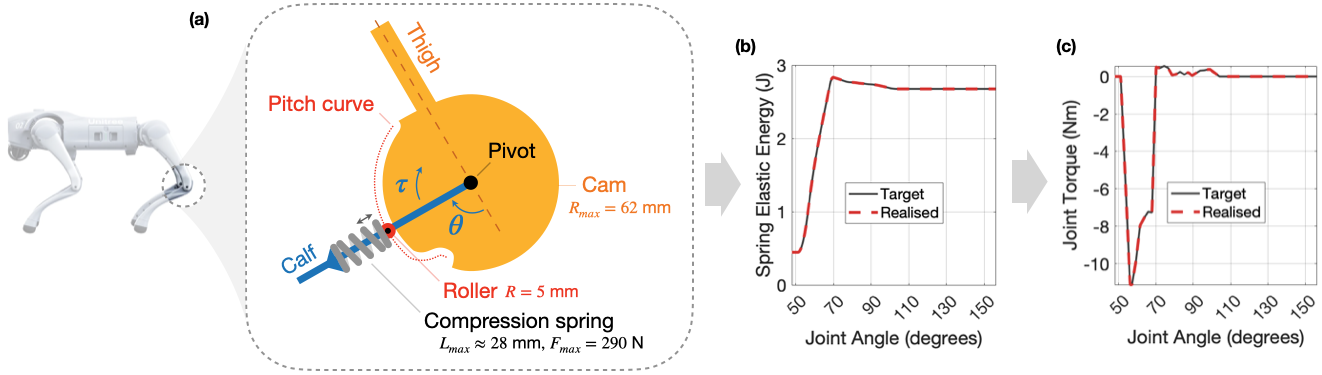}
\caption{\textbf{Conceptual design of a manufacturable PEJ, shown for the rear-calf joint group (the most demanding of the four).} (a) A cam fixed to the joint shaft bears on a roller-tipped, spring-loaded linear follower; as the joint rotates, the cam geometry shapes the deflection of a single preloaded compression spring, producing the prescribed nonlinear elastic profile. Dimensions shown are for the rear-calf design. (b) Target versus realised spring elastic energy as a function of joint angle. (c) Target versus realised joint torque as a function of joint angle. In both (b) and (c) the realised profiles, obtained with a single off-the-shelf steel compression spring, reproduce the distilled targets with negligible error, confirming physical realisability. Full design parameters are given in Appendix~\ref{sec:appendix_camdesign}.}
\label{fig:pej}
\end{figure*}

Although the present study is based on simulation, the distilled PEJs are physically manufacturable. Distillation can produce an arbitrary piecewise-linear torque response, and the PEJ's rotational energy landscape (the torque profile integrated over the joint angle) is one degree smoother---continuous and once-differentiable---and can therefore be realised using the cam principle that we previously demonstrated in the Elastic Rolling Cam~\cite{wu2025encoding}, as well as in multiple other studies~\cite{bidgoly2016design,torrealba2017design,migliore2007novel,wolf2011dlr,ning2021design,baek2022new,arakelian2016gravity}. As shown in Fig.~\ref{fig:pej}(a), a cam fixed to the joint shaft shapes the deflection of a preloaded spring as the joint rotates, so that the spring's stored strain energy $U(\theta)$---and hence the elastic torque $\tau(\theta)=-\mathrm{d}U/\mathrm{d}\theta$---follows a prescribed nonlinear profile set by the cam geometry. There is a direct mapping from the obtained profile through the distillation process and the cam design. In this manufacturability analysis, we adopt a mathematically simple design: a single cam bearing on a roller-tipped, spring-loaded linear follower (Fig.~\ref{fig:pej}(a)). The target torque $\tau_\mathrm{t}(\theta)$ maps to the required energy $U(\theta)=U_0-\int\tau_\mathrm{t}\,\mathrm{d}\theta$ and hence, to the cam pitch radius $r(\theta)=R_b+\sqrt{2U(\theta)/k}-\delta_\mathrm{pre}$, where $R_b$ is the cam's base-circle radius that is $50$~mm in the present design, $k$ is the spring rate, and $\delta_\mathrm{pre}$ the spring's preload deflection. To provide a concrete example, we detail a full design for the most demanding of the four joint groups---the rear calf---in Appendix~\ref{sec:appendix_camdesign}, confirming that the distilled profiles are realisable with off-the-shelf components of acceptable dimensions and mass.

\section{Discussions \& Conclusion}
\label{sec:conclusion}

We introduced Physical Imitation Learning (PIL), a generalisable controller-body co-design framework to outsource active control to the robot body's passive dynamic response, which is designed to replicate a distilled control policy. Specifically, PIL redistributes the actuation effort of a control policy trained via Reinforcement Learning (RL), by decomposing it into a passive and an active component. The passive component is static joint angle--torque profiles distilled from the learned policy, which is implemented as Parallel Elastic Joints (PEJs) that operate in analogy to passive exoskeletons. The active component is the residual policy after distillation, which commands the joints' motors. This allows a significant portion of the actuation effort to be offloaded to the passive PEJs, while the robot still fully reproduces the behaviour of the learned policy with combined actuation from motors and PEJs. Because the active and passive components are both obtained from a single parent policy through post-training distillation, training that policy effectively co-designs both, whilst the policy training itself can follow a conventional RL protocol. It thereby avoids the curse of dimensionality of conventional brain-body co-design approaches that require both body and controller design spaces to be searched simultaneously~\cite{burden2024animals,goff2024investigation}.

In simulation, we applied the PIL pipeline to a Unitree Go2 robot, achieving a 95\% power offload percentage on flat terrain and 13\% on the most difficult terrain the learned policy could cross. The difference in energy savings across terrain levels indicates that difficult terrains require higher actuation margin, while flat terrain allows regular, periodic gaits that can be more readily imitated by the passive PEJs---echoing the principle of passive dynamic walkers. Cross-terrain evaluation demonstrates that terrain-specific policies generalise to lower difficulty levels, though with reduced power offloading. More robust cross-terrain performance combining high offloading on flat terrain with reliable locomotion on difficult terrain is potentially achievable, supported by the broadly similar PEJ responses trained for different terrain levels (Fig.~\ref{fig:pej_curves}(d)). Although the present study is conducted entirely in simulation, the PEJ profiles are manufacturable using the Elastic Rolling Cam~\cite{wu2025encoding} or other existing cam-based mechanical strategies~\cite{bidgoly2016design,torrealba2017design,migliore2007novel,arakelian2016gravity,baek2022new,ning2021design,hu2020novel,medina2016design,tsagarakis2011new,wolf2011dlr}.

Future work will focus on realising robust cross-terrain performance through more advanced training methods, as well as hardware validation on a physical Unitree Go2 robot. We will also explore the generalisability of the PIL framework, which can be applied to a wide range of robots with joint-based morphologies, including humanoids and robot arms. We anticipate PIL to enable future robots whose bodies assist specific control tasks, realising Physical Intelligence without re-inventing control.

\appendix
\section*{APPENDIX}
\setcounter{subsection}{0}
\renewcommand{\thesubsection}{\Alph{subsection}}

\subsection{Prediction of Power Consumption}
\label{sec:power}
Throughout the paper, the motor power consumption is predicted under a zero-regeneration model of mechanical power. Typically, electric actuators in current quadruped hardware dissipate rather than recover energy during negative-work phases, which makes the positive-only motor power a conservative yet realistic proxy for electrical energy consumption. Accordingly, only intervals during which a motor actively performs positive work are counted, whereas intervals of negative work (during which the motor would absorb energy) are set to zero. The instantaneous motor power consumption is
\begin{equation}\label{eq:positive_power}
P(t) = \max\,\bigl(0,\;\tau(t)\,\dot{q}(t)\bigr),
\end{equation}
where $\tau$ and $\dot{q}$ are the motor torque and joint angular velocity. Joule heating in the motor windings and mechanical friction losses are not modelled; only the mechanical power delivered at the joint output is considered.

Locomotion quality is quantified by the mean absolute velocity tracking error,
\begin{equation}\label{eq:tracking_error}
e_v = \frac{1}{T}\int_0^T \bigl\lvert v_{\text{cmd}}(t) - v_{\text{actual}}(t)\bigr\rvert\,dt,
\end{equation}
where $v_{\text{cmd}}(t)$ is the target forward speed, $v_{\text{actual}}(t)$ is the
robot's actual forward speed, and $T=6$\,s is the evaluation duration. A smaller
$e_v$ indicates better velocity tracking. The per-policy, per-terrain tracking
errors are reported in Fig.~\ref{fig:tracking_heatmap}.

\subsection{Reward Weights and Training Configuration}
\label{sec:appendix_weights}

Table~\ref{tab:reward_weights} lists the reward weights used during continued training (Stage~2), which follows the default locomotion-RL setup. The velocity-tracking terms use an exponential kernel: instead of penalising the tracking error directly, the reward is computed as $\exp(-e^2/\text{std}^2)$, which equals $1$ when the robot exactly matches its commanded velocity and decays smoothly toward $0$ as the error $e$ grows. The bandwidth parameter $\text{std}^2$ is set to $0.25$. The base-height reward is a one-sided penalty that activates only when the torso falls below the Go2's nominal standing height of $0.34$\,m, which discourages the undesirable crawling gait that the CoT penalty would otherwise encourage on flat terrain. Both the reference and co-design policies use these same weights, differing only in the CoT weight $\alpha$ (Table~\ref{tab:cot_weights}).

\begin{table}[h]
\centering
\small
\caption{Reward weights for continued training (Stage~2). CoT weights are in Table~\ref{tab:cot_weights}.}
\label{tab:reward_weights}
\setlength{\tabcolsep}{4pt}
\begin{tabular}{ll}
\toprule
\textbf{Reward term} & \textbf{Weight} \\
\midrule
\multicolumn{2}{l}{\textit{Velocity tracking}}\\
\quad\texttt{track\_lin\_vel\_xy\_exp} & $+2.5$ \\
\quad\texttt{track\_ang\_vel\_z\_exp} & $+0.5$ \\
\midrule
\multicolumn{2}{l}{\textit{Body stability}}\\
\quad\texttt{lin\_vel\_z\_l2} & $-2.0$ \\
\quad\texttt{ang\_vel\_xy\_l2} & $-0.05$ \\
\quad\texttt{base\_height\_above\_floor\_exp} & $+0.5$ \\
\quad\texttt{roll\_orientation\_exp} & $+0.36$ \\
\quad\texttt{pitch\_orientation\_exp} & $+0.09$ \\
\midrule
\multicolumn{2}{l}{\textit{Joint / action smoothness}}\\
\quad\texttt{dof\_torques\_l2} & $-$2e-4 \\
\quad\texttt{dof\_acc\_l2} & $-$2.5e-7 \\
\quad\texttt{action\_rate\_l2} & $-0.01$ \\
\midrule
\multicolumn{2}{l}{\textit{Contact / gait}}\\
\quad\texttt{feet\_air\_time} & $+0.01$ \\
\bottomrule
\end{tabular}
\end{table}

\begin{table}[h]
\centering
\caption{CoT weight $\alpha$ for each terrain level.}
\label{tab:cot_weights}
\small
\setlength{\tabcolsep}{3pt}
\begin{tabular}{l@{\hspace{6pt}}c@{\hspace{4pt}}c@{\hspace{4pt}}c@{\hspace{4pt}}c@{\hspace{4pt}}c@{\hspace{4pt}}c@{\hspace{4pt}}c}
\toprule
& \textbf{Flat} & \textbf{L1} & \textbf{L2} & \textbf{L3} & \textbf{L4} & \textbf{L5} & \textbf{L6} \\
\midrule
Co-design & 2.0 & 1.0 & 0.9 & 0.5 & 0.4 & 0.2 & 0.1 \\
Reference & 0.6 & 0.6 & 0.5 & 0.4 & 0.3 & 0.15 & 0.1 \\
\bottomrule
\end{tabular}
\end{table}

\begin{figure}[h]
\centering
\includegraphics[width=\columnwidth]{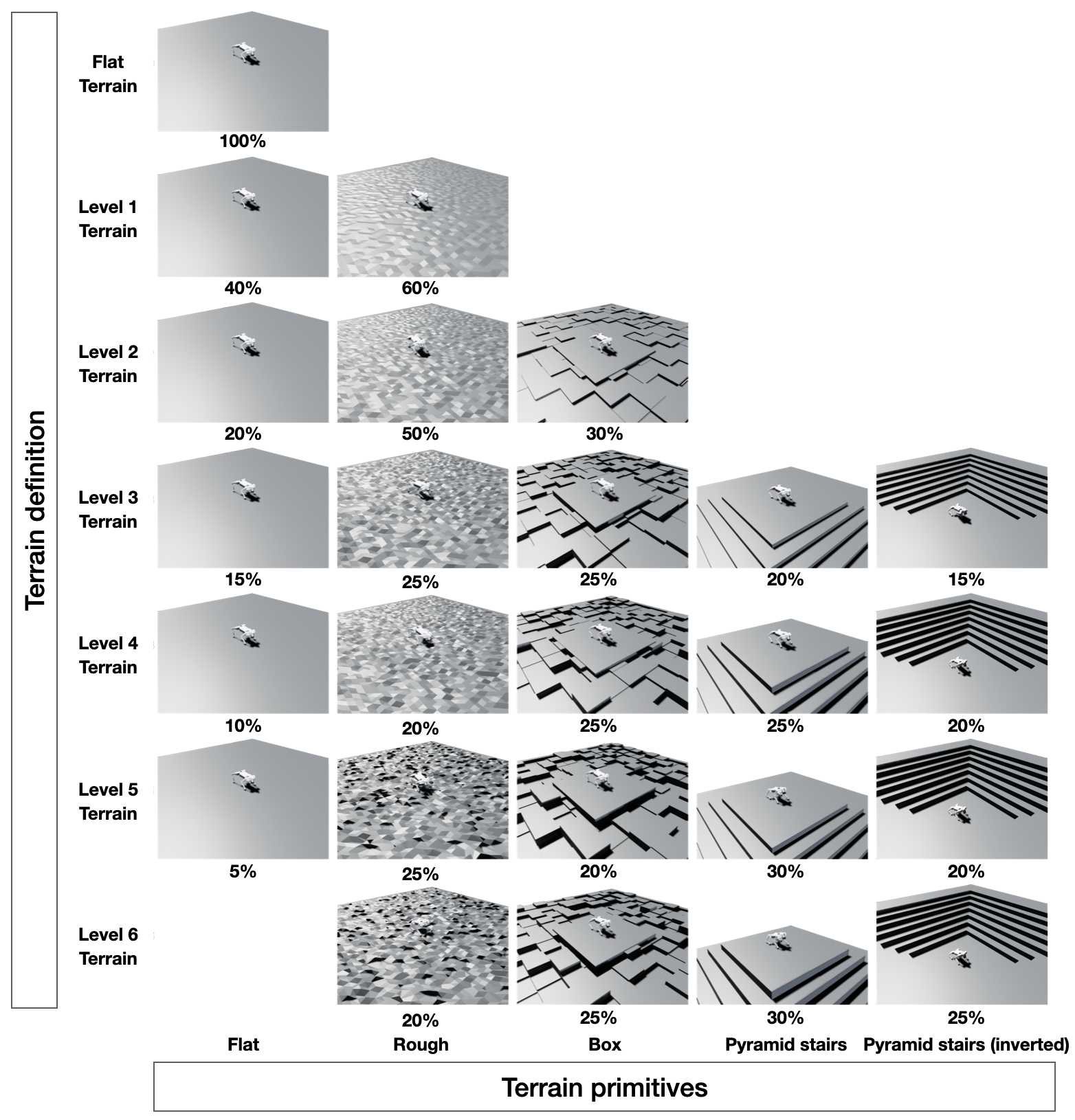}
\caption{\textbf{Terrain types and difficulty levels used in simulation.} Each level is a mixture of five procedurally generated primitives, with the percentage of each type indicated. As the level increases, harder primitives occupy a larger fraction and their dimensions grow.}
\label{fig:terrains}
\end{figure}

The seven terrain difficulty levels used throughout training and evaluation are illustrated in Fig.~\ref{fig:terrains}, with their full geometric specifications listed in Table~\ref{tab:terrain_specs}.

\begin{table}[htbp]
\centering
\small
\caption{Terrain specifications per difficulty level. All heights are in centimetres.}
\label{tab:terrain_specs}
\setlength{\tabcolsep}{4pt}
\begin{tabular}{lccc}
\toprule
\textbf{Level} & \textbf{Random rough} & \textbf{Boxes} & \textbf{Stairs} \\
\midrule
Flat & --- & --- & --- \\
1 & 2.0 -- 2.9 & --- & --- \\
2 & 2.9 -- 3.8 & 6.7 -- 8.3 & --- \\
3 & 3.8 -- 4.7 & 8.3 -- 10.0 & --- \\
4 & 4.7 -- 5.6 & 10.0 -- 11.7 & 11 -- 13 \\
5 & 5.6 -- 6.4 & 11.7 -- 13.3 & 13 -- 15 \\
6 & 6.4 -- 7.3 & 13.3 -- 15.0 & 15 -- 17 \\
\bottomrule
\end{tabular}
\end{table}

\subsection{Terrain-Specific Power Consumption}
\label{sec:appendix_power}

Table~\ref{tab:power_data} reports the average motor power consumption underlying the offload percentages in Fig.~\ref{fig:combined_results}. For each policy and terrain level, the ``before'' value is the total motor power demanded when the motor supplies the full joint torque, and the ``after'' value is the residual active power once the distilled PEJ torque is subtracted. The offload percentage is obtained from these two values via Eq.~\eqref{eq:offload_ratio}. 

\begin{table}[h]
\centering
\small
\caption{Average motor power consumption (in watts) before and after PEJ assistance, the power consumption includes the hip motors.}
\label{tab:power_data}
\setlength{\tabcolsep}{4pt}
\begin{tabular}{lccccccc}
\toprule
 & \textbf{Flat} & \textbf{L1} & \textbf{L2} & \textbf{L3} & \textbf{L4} & \textbf{L5} & \textbf{L6} \\
\midrule
\multicolumn{8}{l}{\textit{Co-design policy}}\\
\quad Before (W) & 21.7 & 19.6 & 24.3 & 26.8 & 31.0 & 39.4 & 49.5 \\
\quad After (W)  & 1.12 & 3.70 & 8.65 & 16.6 & 22.3 & 32.3 & 43.0 \\
\quad Offload (\%) & 94.8 & 81.2 & 64.4 & 38.1 & 28.0 & 18.2 & 13.1 \\
\midrule
\multicolumn{8}{l}{\textit{Reference policy}}\\
\quad Before (W) & 18.5 & 21.3 & 22.6 & 25.4 & 29.8 & 38.2 & 48.6 \\
\quad After (W)  & 15.4 & 17.8 & 18.5 & 22.4 & 25.2 & 33.0 & 43.0 \\
\quad Offload (\%) & 16.6 & 16.3 & 18.1 & 11.7 & 15.4 & 13.6 & 11.6 \\
\bottomrule
\end{tabular}
\end{table}

\subsection{Cam Design for the Rear-Calf PEJ}
\label{sec:appendix_camdesign}

As a concrete demonstration of manufacturability (Section~\ref{sec:PEJ_feasibility}), we detail the design of the rear-calf PEJ, the most demanding of the four joint groups: its distilled target profile requires the largest elastic energy variation ($\Delta U = 2.4$~J) and the largest peak torque ($11.1$~N\,m). The profile is defined over the joint's full range of motion ($48^\circ$--$156^\circ$).

The profile is realised with a single off-the-shelf steel compression spring (Lee Spring LC112L01M: spring rate $k = 14.9$~N\,mm$^{-1}$, free length $38$~mm, maximum compression $19.5$~mm, mass ${\approx}\,22$~g). The mechanism is dimensioned so that the spring reaches full compression---a peak force of $290$~N---at the maximum-energy state. Following the cam mapping of Section~\ref{sec:PEJ_feasibility}, the resulting cam has a base-circle radius $R_b = 50$~mm and a maximum outer radius of $62$~mm, and reproduces the target torque and stored-energy profiles with negligible error (Fig.~\ref{fig:pej}(b,c)). This confirms that the distilled PEJ torque--angle profiles are physically realisable with components of acceptable dimensions and mass. The design programme will be provided in our online repository.

\section*{Acknowledgment}
This work was supported by the Royal Society Newton International Fellowship NIF/R1/241883, and the Engineering and Physical Sciences Research Council (EPSRC) grant EP/S021795/1.

\bibliographystyle{IEEEtran}
\bibliography{references}

\EOD

\end{document}